\documentclass[a4paper]{article}
\usepackage{INTERSPEECH2016}

\usepackage{graphicx}
\usepackage{amssymb,amsmath,bm}
\usepackage{textcomp}
\usepackage{mathtools}

\ninept

\title{Towards End-to-End Speech Recognition with Deep Convolutional Neural Networks}

\makeatletter
\def\name#1{\gdef\@name{#1\\}}
\makeatother \name{{\em Ying Zhang, Mohammad Pezeshki, Phil\'{e}mon Brakel, Saizheng Zhang, C\'{e}sar Laurent}\\
{\em Yoshua Bengio$^{1}$, Aaron Courville$^{2}$}}
\address{Departement d'informatique et de recherche op\'{e}rationnelle \\
  Universit\'{e} de Montr\'{e}al, Montr\'{e}al, QC H3C 3J7 \\
$^1$ CIFAR Senior Fellow, $^2$ CIFAR Fellow\\
\small \tt \{zhangy, mohammad.pezeshki, philemon.brakel, saizheng.zhang, cesar.laurent\}@umontreal.ca\\
\small \tt \{Yoshua.Bengio, Aaron.Courville\}@umontreal.ca}

\begin{document}

  \maketitle
  \begin{abstract}
    Convolutional Neural Networks (CNNs) are effective models for reducing spectral
    variations and modeling spectral correlations in acoustic features
    for automatic speech recognition (ASR). Hybrid speech recognition systems
    incorporating CNNs with Hidden Markov Models/Gaussian Mixture Models (HMMs/GMMs)
    have achieved the state-of-the-art in various benchmarks. Meanwhile,
    Connectionist Temporal
    Classification (CTC) with Recurrent Neural Networks (RNNs), which is proposed
    for labeling unsegmented sequences, makes it feasible to train an `end-to-end'
    speech recognition system instead of hybrid settings. However, RNNs are 
    computationally expensive and sometimes difficult to train. In this paper, inspired by
    the advantages of both CNNs and the CTC approach, we propose an end-to-end speech
    framework for sequence labeling, by combining hierarchical CNNs with CTC
    directly without recurrent connections. By evaluating the approach on the TIMIT
    phoneme recognition task, we show that
    the proposed model is not only computationally efficient, but also
    competitive with the existing baseline systems. Moreover, we argue that CNNs
    have the capability to model temporal correlations with appropriate context
    information.
  \end{abstract}
  
  \noindent{\bf Index Terms}: speech recognition, convolutional neural networks,
  connectionist temporal classification

  \section{Introduction}
    Recently, Convolutional Neural Networks (CNNs) \cite{lecun1998gradient} have
    achieved great success in acoustic modeling \cite{abdel2012applying,
    sainath2013deep, sainath2013improvements}. In the context of Automatic Speech Recognition, CNNs are usually
    combined with HMMs/GMMs \cite{mohamed2012acoustic, hinton2012deep}, like regular Deep Neural Networks (DNNs),
    which results in a hybrid system \cite{abdel2012applying, sainath2013deep, sainath2013improvements}. In the typical hybrid system, the neural net is trained to predict frame-level targets obtained from a forced alignment generated by an HMM/GMM system. The temporal modeling and decoding operations are still handled by an HMM but the posterior state predictions are generated using the neural network.
    
%
    This hybrid approach is problematic in that training the different modules separately with different criteria may not be optimal for solving the final task. As a consequence, it often requires additional hyperparameter tuning for each training stage which can be laborious and time consuming. Furthermore, these issues have motivated a recent surge of interests in training `end-to-end' systems \cite{hannun2014deep,bahdanau2015end,miao2015eesen}. End-to-end neural systems for speech recognition typically replace the HMM with a neural network that provides a distribution over sequences directly. Two popular neural network sequence models are Connectionist Temporal Classification (CTC) \cite{graves2006connectionist} and recurrent models for sequence generation \cite{bahdanau2015end,chorowski2015attention}.
    
	To the best of our knowledge, all end-to-end neural speech recognition systems employ recurrent neural networks in at least some part of the processing pipeline. The most successful recurrent neural network architecture used in this context is the Long Short-Term Memory (LSTM) \cite{graves2013speech,graves2012sequence,
    hochreiter1997long,vinyals2012revisiting}.
    For example, 
    a model with multiple layers of bi-directional LSTMs and CTC on top which is pre-trained with the transducer networks \cite{graves2013speech,graves2012sequence} obtained the state-of-the-art on the TIMIT dataset.     
%
%
After these successes on phoneme recognition, similar systems have been proposed in which multiple layers of RNNs were combined with CTC to perform large vocabulary continuous speech recognition \cite{hannun2014deep,miao2013deep}. It seems that RNNs have become somewhat of a default method for end-to-end models while hybrid systems still tend to rely on feed-forward architectures.

While the results of these RNN-based end-to-end systems are impressive,  there are two important
    downsides to using RNNs/LSTMs: (1) The training speed can be very slow due
    to the iterative multiplications over time when the input sequence is
    very long; (2) The training process is sometimes tricky due to the well-known problem of gradient vanishing/exploding \cite{hochreiter1991untersuchungen,
    bengio1994learning}. Although various approaches have been proposed to address
    these issues, such as data/model parallelization across multiple
    GPUs \cite{hannun2014deep, sutskever2014sequence}
    and careful initializations for recurrent connections \cite{le2015simple},
    those models still suffer from computationally intensive and otherwise demanding training procedures. 
    
    Inspired by the strengths of both CNNs and CTC, we propose an
    end-to-end speech framework in which we combine CNNs with CTC without
    intermediate recurrent layers. We present experiments on the TIMIT dataset and show that such a system is able to obtain results that are comparable to those obtained with multiple layers of LSTMs.
    The only previous attempt to combine CNNs with CTC that we know about \cite{songend}, led to results that were far from the state-of-the-art. It is not straightforward to incorporate CNN into an end-to-end manner since the task may require the model to incorporate long-term dependencies. While RNNs can learn these kind of dependencies and have been combined with CTC for this very reason, it was not known whether CNNs were able to learn the required temporal relationships. 
    
    In this paper, we argue that in a CNN of sufficient depth, the higher-layer features are capable
    of capturing temporal dependencies with suitable
    context information. Using small filter sizes along the spectrogram frequency axis, the model is able to learn fine-grained localized features, while multiple stacked
    convolutional layers help to learn diverse features on different
    time/frequency scales and provide the required non-linear modeling capabilities.
   
    Unlike the time windows applied in DNN systems \cite{abdel2012applying,sainath2013deep,sainath2013improvements},
    the temporal modeling is deployed within convolutional layers, where we perform a 2D
    convolution similar to vision tasks, and multiple convolutional layers are stacked to provide a relatively large context window for each output prediction of the highest layer. The convolutional layers are followed by multiple fully connected layers and, finally, CTC is added on the top of the model. 
    Following the suggestion from \cite{sainath2013improvements}, we only
    perform pooling along the frequency band on the first convolutional layer. Specifically, we evaluate our model on phoneme recognition for the TIMIT dataset.
 
 \begin{figure}[t]
  \centering
  \includegraphics[width=\linewidth]{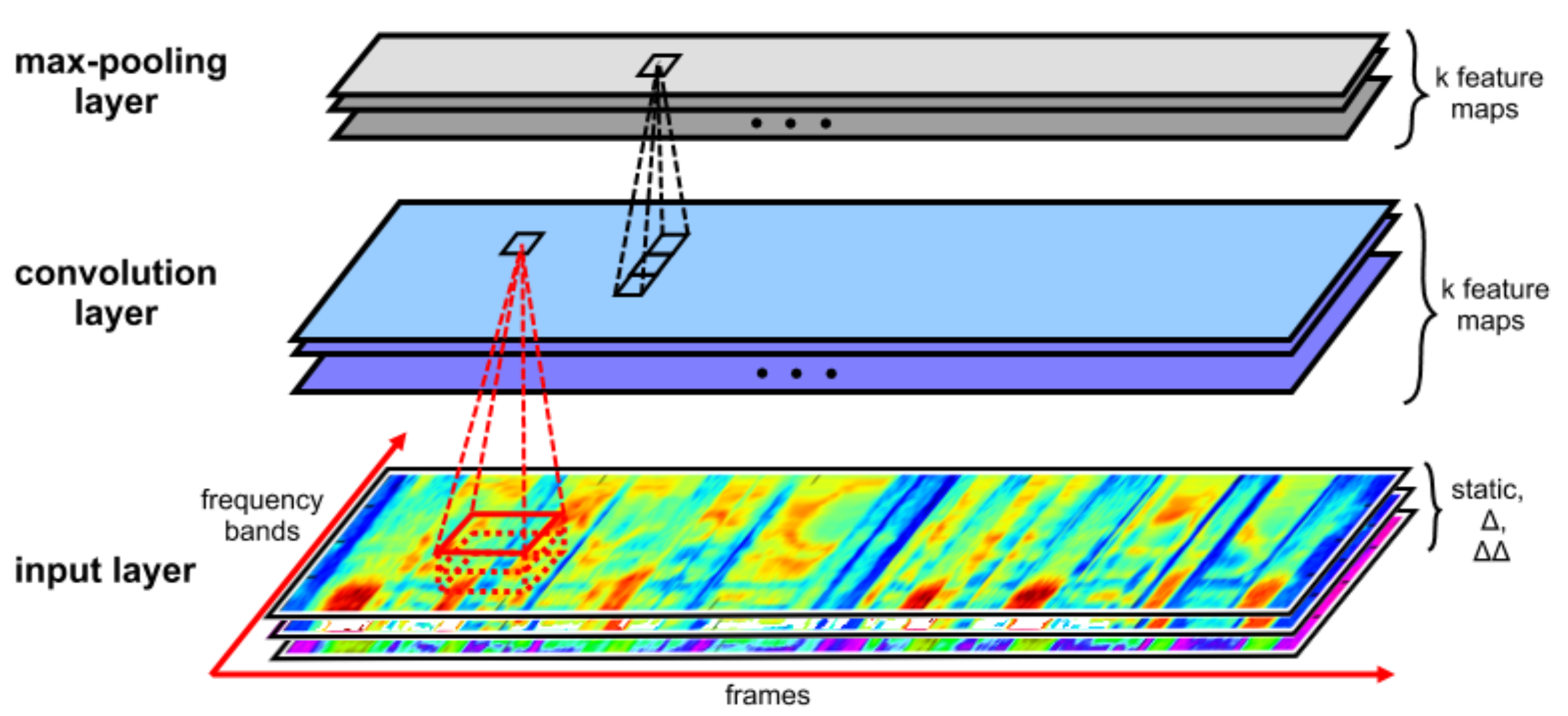}
  \caption{{\it The convolution layer and max-pooling layer applied upon input
  features.}}
  \label{fig:conv_pooling}
\end{figure}
    
 \section{Convolutional Neural Networks}
 	Most of the CNN  models \cite{abdel2012applying,sainath2013deep,sainath2013improvements}
    in the speech domain have large filters and use
    limited weight sharing which splits the features into limited frequency bands while performing convolution separately and the convolution is usually applied with no more than 3 layers.
    In this section, we describe our CNN acoustic model
    whose architecture
    is different from the above. The complete CNN includes stacked
    convolutional and pooling layers, at the top of which are multiple
    fully-connected layers. 

    Since CNNs are adept at modeling local structures in the inputs,
we use log mel-filter-bank (plus energy term) coefficients with deltas and
delta-deltas which preserve the local correlations of the spectrogram.

    \subsection{Convolution}
    As shown in Figure~\ref{fig:conv_pooling}, given a sequence of acoustic feature values $\mathbf{X}\in \mathbb{R}^{c\times b\times f}$ with number of channels $c$,
    frequency bandwidth $b$, and time length $f$, 
    the convolutional layer convolves $\mathbf{X}$ with $k$ filters $\{\mathbf{W}_i\}_k$ where
    each $\mathbf{W}_i\in\mathbb{R}^{c\times m\times n}$ is a $3D$ tensor with its width along the frequency axis equal to $m$ and its length along frame axis equal to $n$.
    The resulting $k$ pre-activation feature maps consist of  a $3$D tensor $\mathbf{H}\in\mathbb{R}^{k\times b_H\times f_H}$, in which each feature map $\mathbf{H}_i$ is computed as follows:
    	\begin{equation}
          \mathbf{H}_i = \mathbf{W}_i * \mathbf{X} + b_{i}, \ \ \ i=1, \cdots, k.
          \label{eq1}
        \end{equation}
    The symbol $*$ denotes the convolution operation and $b_i$ is a bias parameter. There are three points that are worth mentioning:\break 
    (1) The sequence length $f_H$ of $H$ 
    after convolution is guaranteed to be equal to the input $\mathbf{X}$'s sequence length $f$ by applying zero padding along the frame axis before each convolution; (2) The convolution stride is chosen to be $1$ for all the convolution operations in our model; (3) We do not use \textit{limited weight sharing} which splits the frequency bands into groups of limited bandwidths and convolution is done within
    each group separately. Instead, we perform the convolution over $\mathbf{X}$ not only along the frequency axis but also along the time axis, which results in a simple 2D convolution commonly used in computer vision.
    
	\subsection{Activation Function}
    The pre-activation feature maps $\mathbf{H}$ are passed through non-linear activation functions. We introduce three activation functions in the following and show their functionalities in the convolutional layer as an example, notice that all the operations below are element-wise.
    
    \subsubsection{Rectifier Linear Unit}
    Rectifier Linear Unit (ReLU) \cite{glorot2011deep} is a piece-wise linear activation function that outputs zero if the input is negative and outputs the input itself otherwise. Formally, given single feature map $\mathbf{H}_i$, a ReLU function is defined as follows:
    \begin{equation}
    \tilde{\mathbf{H}}_i = \textit{max}(0, \mathbf{H}_i),
    \end{equation}  
in which $\mathbf{H}$ and $\tilde{\mathbf{H}}$ are the input and output respectively.
    
    \subsubsection{Parametric Rectifier Linear Unit}
    The Parametric Rectifier Linear Unit (PReLU) \cite{he2015delving} is an extension of the ReLU in which the output of the model in the regions that input is negative is a linear function of the input with a slope of $\alpha$. PReLU is formalized as:
    \begin{equation}
     \tilde{\mathbf{H}}_i=
    \begin{cases}
      \mathbf{H}_i, & \text{if}\ \mathbf{H}_i>0 \\
      \alpha \mathbf{H}_i, & \text{otherwise}
    \end{cases}
    \end{equation}
    The extra parameter $\alpha$ is usually initialized to 0.1 and can be trained using backpropagation.  
\begin{figure}[t]
  \centering
  \includegraphics[width=210pt]{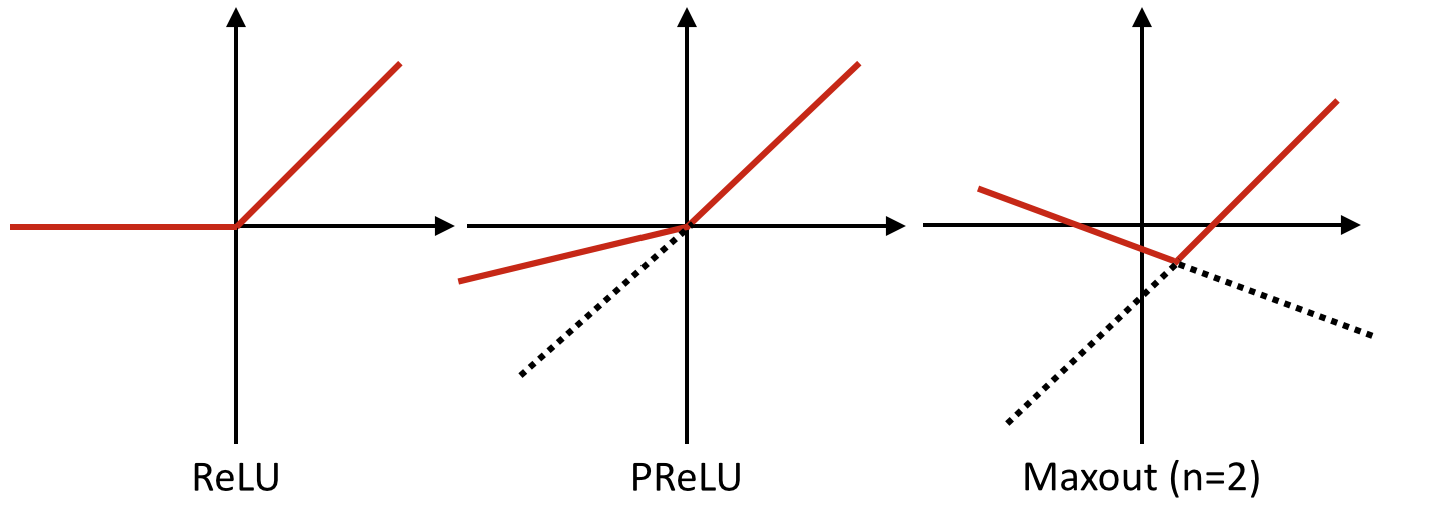}
  \caption{{\it ReLU, PReLU and Maxout activation functions.}}
  \label{fig:activations}
\end{figure}
    \subsubsection{Maxout}
    Another type of activation function which has been shown to improve the results for the task of speech recognition \cite{miao2013deep,cai2013deep,zhang2014improving,toth2015phone} is the \textit{maxout} function \cite{goodfellow2013maxout}.
    Following the same computational process as in \cite{goodfellow2013maxout}, we take the
    number of piece-wise linear functions as $2$ for example. Then for $\tilde{\mathbf{H}}$ we have:
    \begin{equation}
    \tilde{\mathbf{H}}_i = \textit{max}(\mathbf{H}'_i, \mathbf{H}''_i),
    \end{equation}
    where for $\mathbf{H}'_i$ and $\mathbf{H}''_i$ we have:
    \begin{eqnarray}
    \mathbf{H}'_{i} = \mathbf{W}'_{i} * \mathbf{X} + b'_{i},\ \ \ \mathbf{H}''_{i} = \mathbf{W}''_{i} * \mathbf{X} + b''_{i}, 
    \end{eqnarray}
    which are two linear feature map candidates after the convolution, and $\mathbf{X}$ is the input of the convolutional layer at $\mathbf{H}_i$. Figure \ref{fig:activations} depicts the ReLU, PReLU, and Maxout activation functions.

    \subsection{Pooling}
\begin{figure}[t]
        \centering
        \includegraphics[width=130pt]{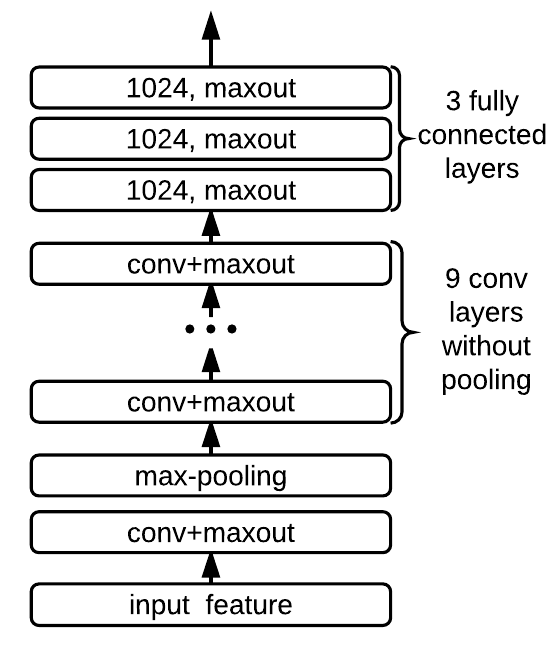}
        \caption{{\it Network structure for phoneme recognition on the TIMIT dataset. The model consists of 10 convolutional layers followed by 3 fully-connected layers on the top. All convolutional layers have the filter size of $3\times 5$ and we use max-pooling with size of $3\times 1$ only after the first convolutional layer. First and second numbers correspond to frequency and time axes respectively.}}
        \label{fig: Convolutional Structure}
      \end{figure}

    After the element-wise non-linearities, the features will pass through a max-pooling layer which outputs the maximum unit from $p$ adjacent units. We do pooling only along the frequency axis since it helps to reduce spectral variations within the same speaker and between
    different speakers \cite{lecun1995convolutional}, while pooling in time has been shown to be less helpful \cite{sainath2013improvements}. Specifically, suppose that the $i$ th feature map before and after pooling are $\tilde{\mathbf{H}}_i$ and $\hat{\mathbf{H}}_i$, then $[\hat{\mathbf{H}}_i]_{r,t}$ at position $(r, t)$ is computed by:
    \begin{equation}
    [\hat{\mathbf{H}}_i]_{r,t} =  
    \textit{max}^p_{j=1}\{[\tilde{\mathbf{H}}_i]_{r\times s + j, t}\},
    \end{equation}
    where $s$ is the step size and $p$ is the pooling size, and all the $[\tilde{\mathbf{H}}_i]_{r\times s + j, t}$ values inside the \textit{max} have the same time index $t$. Consequently, the feature maps after pooling have the same sequence lengths as the ones before pooling. As shown in Figure~\ref{fig: Convolutional Structure}, we follow the suggestions from \cite{sainath2013improvements} that the max pooling is performed only once after the first convolutional layer. Our intuition is that as more pooling layers are applied, units in higher layers would be less discriminative with respect to the variations in input features.

\section{Connectionist Temporal Classification}
Consider any sequence to sequence mapping task in which $\textbf{X} = \{X_1, ..., X_T\}$ is the input sequence and $\textbf{Z} = \{Z_1, \cdots, Z_{L}\}$ is the target sequence. In the case of speech recognition, $\textbf{X}$ is the acoustic signal and $\textbf{Z}$ is a sequence of symbols. In order to train the neural acoustic model, $Pr(\textbf{Z}|\textbf{X})$ must be maximized for each input-output pair.

One way to provide a distribution over variable length output sequences given some much longer input sequence, is to introduce a many-to-one mapping of latent variable sequences $\mathbf{O}=\{O_1, \cdots,O_T\}$ to shorter sequences that serve as the final predictions. The probability of some sequence $\textbf{Z}$ can then be defined to be the sum of the probabilities of all the latent sequences that map to that sequence.
Connectionist Temporal Classification (CTC) \cite{graves2012supervised} specifies a distribution over latent sequences by applying a softmax function to the output of the network for every time step, which provides a probability for emitting each label from the alphabet of output symbols at that time step $Pr(O_t|\mathbf{X})$. An extra \emph{blank} output class `-' is introduced to the alphabet for the latent sequences to represent the probability of not outputting a symbol at a particular time step. Each latent sequence sampled from this distribution can now be transformed into an output sequence using the many-to-one mapping function $\sigma(\cdot)$ which first merges the repetitions of consecutive non-blank labels to a single label and subsequently removes the blank labels as shown in Equation \ref{sigma}:
\begin{equation}
\left.
\begin{array}{l}
\sigma(a,b,c,-,-)\\
\sigma(a,b,-,c,c)\\
\sigma(a,a,b,b,c)\\
\sigma(-,a,-,b,c)\\
\vdotswithin{\ldots}\\
\sigma(-,-,a,b,c)
\end{array}\right\}=(a,b,c).
\label{sigma}
\end{equation} 
Therefore, the final output sequence probability is a summation over all possible sequences $\pi$ that yield to $\textbf{Z}$ after applying the function $\sigma$:
\begin{equation}
Pr(\textbf{Z}|\textbf{X}) = \Sigma_{\mathbf{o} \in \sigma^{-1}(\textbf{Z})} Pr(\mathbf{O}|\textbf{X}).
\label{ctc}
\end{equation}
A dynamic programming algorithm similar to the forward algorithm for HMMs \cite{graves2012supervised} is used to compute the sum in Equation \ref{ctc} in an efficient way. The intermediate values of this dynamic programming can also be used to compute the gradient of $\ln Pr(\mathbf{Z}|\mathbf{X})$ with respect to the neural network outputs efficiently.

To generate predictions from a trained model using CTC, we use the \textit{best path decoding} algorithm. Since the model assumes that the latent symbols are independent given the network outputs in the framewise case, the latent sequence with the highest probability is simply obtained by emitting the most probable label at each time-step. The predicted sequence is then given by applying $\sigma(\cdot)$ to that latent sequence prediction:
\begin{equation}
\textbf{L} \approx \sigma(\pi^*),
\end{equation}
in which $\pi^*$ is the concatenation of the most probable output and is formalized by $\pi^* = \mathrm{Argmax}_\pi Pr(\pi|\textbf{X})$. Note that this is not necessarily the output sequence with the highest probability. Finding this sequence is generally not tractable and requires some approximate search procedure like a beam-search.

\section{Experiments}
	In this section, we evaluate the proposed model on phoneme recognition for the TIMIT dataset. The model architecture is shown in Figure~\ref{fig: Convolutional Structure}.  

	\subsection{Data}
    We evaluate our models on the TIMIT \cite{garofolo1993darpa} corpus where we use the standard 462-speaker training set
    with all SA records removed. The 50-speaker development set is used for early stopping. The evaluation is
performed on the core test set (including 192 sentences). The raw audio is transformed into 40-dimensional log mel-filter-bank (plus energy term) coefficients with deltas and
delta-deltas, which results in 123 dimensional features. Each dimension is normalized to have zero mean and unit
variance over the training set. We use 61 phone labels plus a blank label for training and then the output is mapped to 39 phonemes for scoring.
	
   \subsection{Models}
   Our best model consists of 10 convolutional layers and 3 fully-connected hidden
    layers. Unlike the other layers, the first convolutional layer is followed by a pooling layer, which is described in section 2. The pooling size is $3\times 1$, which means we only pool over the frequency axis. The filter size is $3\times 5$ across the layers. The model has 128 feature maps in the first four convolutional layers and 256 feature maps in the remaining six convolutional layers. Each fully-connected layer has 1024 units. Maxout with 2 piece-wise linear functions is used as the activation function. Some other architectures are also evaluated for comparison, see section 4.4 for more details.
    
   \subsection{Training and Evaluation}   
    To optimize the model, we use Adam \cite{kingma2014adam} with learning rate $10^{-4}$. Stochastic gradient descent with learning rate $10^{-5}$ is then used for fine-tuning. Batch size 20 is used during training. The
    initial weight values were drawn uniformly from the interval $[-0.05, 0.05]$. Dropout \cite{srivastava2014dropout} with a probability of $0.3$ is added across the layers except for the input and output layers . L2 norm with coefficient $1e-5$ is applied at fine-tuning stage. At test time, simple best path decoding (at the CTC frame level) is used to get the predicted sequences. 
    
    \subsection{Results}
    Our model achieves $18.2\%$ phoneme error rate on the core test set, which is slightly better than the LSTM baseline model and the transducer model with an explicit
RNN language model. The details are presented in Table \ref{tab:per}. Notice that the CNN model could take much less time to train in comparison with the LSTM model when keeping roughly the same number of parameters. In our setup on TIMIT, we get  $2.5\times$ faster training speed by using the CNN model without deliberately optimizing the implementation. We suppose that the gain of the computation efficiency might be more dramatic with a larger dataset.
   
   To further investigate the different structural aspects of our model, we disentangle the analysis into three sub-experiments considering the number of convolutional layers, the filter sizes and the activation functions, as shown in table \ref{tab:per}. It turns out that the model may benefit from (1) more layers, which results in more nonlinearities and larger input receptive fields for units in the top layers; (2) reasonably large context windows, which help the model to capture the spatial/temporal relations of input sequences in reasonable time-scales; (3) the Maxout unit, which has more functional freedoms comparing to ReLU and parametric ReLU.
   
   	\begin{table}[!th]
        \caption{\label{tab:per} {Phoneme Error Rate (PER) on TIMIT. 'NP' is the number of parameters. 'BiLSTM-3L-250H' denotes the model has 3 bidirectional LSTM layers with 250 units in each direction. In the CNN model, $(3, 5)$ is the filter size. Results suggest that deeper architecture and larger filter sizes leads to better performance. The best performing model on Development set, has a test PER of 18.2 \%}}
        \vspace{.1mm}
        \begin{center}
          \begin{tabular}{lccc}
            \hline
            \hline
             Model & NP & Dev PER&Test PER\\
            \hline
              BiLSTM-3L-250H \cite{graves2013speech} & 3.8M &- &$18.6\%$\\
              BiLSTM-5L-250H \cite{graves2013speech} & 6.8M &- &$18.4\%$ \\
              TRANS-3L-250H \cite{graves2013speech} & 4.3M &- &$18.3\%$ \\
            \hline
            CNN-(3,5)-10L-ReLU & 4.3M &$17.4\%$ &$19.3\%$\\
            CNN-(3,5)-10L-PReLU & 4.3M &$17.2\%$ &$18.9\%$\\
              \hline
            CNN-(3,5)-6L-maxout & 4.3M &$18.7\%$ &$21.2\%$\\
            CNN-(3,5)-8L-maxout & 4.3M &$17.7\%$ &$19.8\%$\\
            CNN-(3,3)-10L-maxout & 4.3M &$18.4\%$ &$19.9\%$\\
            CNN-(3,5)-10L-maxout & 4.3M &$\textbf{16.7}\%$ &$\textbf{18.2}\%$\\
          \hline
          \end{tabular}
          \end{center}
      \end{table}

  \section{Discussion}
Our results showed that convolutional architectures with CTC cost can achieve results comparable to the state-of-the-art by adopting the following methodology: (1) Using a significantly deeper architecture that results in a more non-linear function and also wider receptive fields along both frequency and temporal axes; (2) Using maxout non-linearities in order to make the optimization easier; and (3) Careful model regularization that yields better generalization in test time, especially for small datasets such as TIMIT, where over-fitting happens easily.

   
    We conjecture that the convolutional CTC model might be easier to train on phoneme-level sequences rather than the character-level. Our intuition is that the local structures within the phonemes are more robust and can easily be captured by the model. Additionally, phoneme-level training might not require the modeling of many long-term dependencies in comparison with character-level training. As a result, for a convolutional model, learning the phonemes structure seems to be easier, but empirical research needs to be done to investigate if this is indeed the case.
    
    Finally, an important point that favors convolutional over recurrent architectures is the training speed. In a CNN, the training time can be rendered virtually independent of the length of the input sequence due to the parallel nature of convolutional models and the highly optimized CNN libraries available \cite{chetlur2014cudnn}. Computations in a recurrent model are sequential and cannot be easily parallelized. The training time for RNNs increases at least linearly with the length of the input sequence.  
    
  \section{Conclusions}
	 In this work, we present a CNN-based end-to-end speech recognition framework without recurrent neural networks which are widely used in speech recognition tasks. We show promising results on the TIMIT dataset and conclude that the model has the capacity to learn the temporal relations that are required for it to be integrated with CTC. We already observed a gain in computational efficiency on the TIMIT dataset and training the model on large vocabulary datasets and integrate with the language model would be a part of our further study. Another
     interesting direction is to apply Batch Normalization \cite{ioffe2015batch} to the current model.
   
  \section{Acknowledgements}
    The experiments were conducted using Theano \cite{bastien2012theano}, Blocks
and Fuel \cite{van2015blocks}. The authors would like to acknowledge the funding
support from Samsung, NSERC, Calcul Quebec, Compute Canada, the Canada Research Chairs and CIFAR. The authors would like to thank Dmitriy Serdyuk, Dzmitry Bahdanau, Arnaud Bergeron, and Pascal Lamblin for their helpful comments.

  \newpage
  \eightpt
  \bibliographystyle{IEEEtran}

  \bibliography{mybib}


\end{document}